# An Automated Knowledge Mining and Document Classification System with Multi-model Transfer Learning

**Jia Wei Chong[1], ZhiYuan Chen[1*], and Mei Shin Oh[2]**

[1]School of Computer Science, University of Nottingham Malaysia, Jalan Broga, 43500 Semenyih, Selangor, Malaysia

[2]CAD-IT Consultants (M), Jalan SS7/19, Kelana Jaya, 47301 Petaling Jaya, Selangor, Malaysia

[*]Corresponding Author: ZhiYuan Chen. Email: Zhiyuan.Chen@nottingham.edu.my

**Abstract:** Service manual documents are crucial to the engineering company as they provide guidelines and knowledge to service engineers. However, it has become inconvenient and inefficient for service engineers to retrieve specific knowledge from documents due to the complexity of resources. In this research, we propose an automated knowledge mining and document classification system with novel multi-model transfer learning approaches. Particularly, the classification performance of the system has been improved with three effective techniques: fine-tuning, pruning, and multi-model method. The fine-tuning technique optimizes a pre-trained BERT model by adding a feed-forward neural network layer and the pruning technique is used to retrain the BERT model with new data. The multi-model method initializes and trains multiple BERT models to overcome the randomness of data ordering during the fine-tuning process. In the first iteration of the training process, multiple BERT models are being trained simultaneously. The best model is then selected for the next phase of the training process with another two iterations and the training processes for other BERT models will be terminated. The performance of the proposed system has been evaluated by comparing with two robust baseline methods, BERT and BERT-CNN. Experimental results on a widely used Corpus of Linguistic Acceptability (CoLA) dataset have shown that the proposed techniques perform better than these baseline methods in terms of accuracy and MCC score.

# 1 Introduction

A service manual contains much information on the company's product, such as the maintenance, repair, and servicing of the product. These manuals are crucial to the engineering company as they will act as a guideline and knowledgebase for the engineers when they need to work on the product upon requested by the client. However, it has become inconvenient and inefficient for service engineers to retrieve specific knowledge from the service manual since the document typically has complex and different structures. This is also a barrier for a new service engineer to acquire new technical expertise in an effective and timely manner.

In this paper, we propose an automated knowledge mining and document classification system to solve this problem. The idea is to mine data and build a knowledgebase system from service manual documents. With the knowledgebase, service engineers could directly retrieve and gain knowledge without referring back to the original service manual. The proposed system works as an improved version of a diagnostic tool that offers better accuracy in knowledge mining and document classification. There are two main components of the system, namely, data extraction and data classification. For data extraction, the knowledge will be mined from various documents with different structure, such as free text, or in tables. The system will automatically convert the input documents into the format that the classification component can process. Subsequently, data classification using Natural Language Processing (NLP) will be applied. A machine learning model, Bidirectional Encoder Representations from Transformers (BERT) is integrated as a system engine to classify the pre-processed documents into the categories. And then, the classified document is stored into the knowledgebase accordingly based on the symptoms, causes, and possible solutions.

The main focus of this study is to propose a novel multi-model transfer learning approach with pruning and fine-tuning processes, which aims to improve the performance of the pre-trained BERT model in the task of text classification. An additional untrained layer is added to the pre-trained BERT model so that it could be trained and fine-tuned accordingly to the given task. A multi-model BERT is introduced to overcome the randomness of training data in order to achieve the best accuracy with the given dataset. Pruning is introduced to stimulate further learning of the model, and fine-tuning is also used to enable the model to focus on a specific domain task. A web application is also included in this research to host the deployed model and to provide a web interface for user interaction.

## 2 Related Work

### 2.1 Automated Knowledge Mining and Text Classification Systems

Riel and Boonyasopon propose Content Analysis Toolkit (CAT), a text mining tool to mine knowledge by extracting explicit and implicit knowledge from a large collection of electronic text written documents (Riel & Boonyasopon, 2009). Pre-process of data is widely used in NLP tasks for dimensionality reduction by data cleaning and transform the data into an analyzable format. However, pre-processing techniques such as fine-tuning and pruning methods have not been applied. This may lead to an increase of noise in the model due to the use of meaningless data in training, making it impractical to be generic (used by any domain). An expert in the domain knowledge is required to configure and review the system output during the model's training to achieve better accuracy. Sarioglu proposed using NLP features in the medical research community on patient reports to generate a structured output (Sarioglu et al., 2012). In their work, binary topic classification is used to determine the performance of the system. This classification is done by categorizing the topic of the input data into two classes: a positive class which has the correct corresponding medical terms, and a negative class which has words that do not belong to the class. Topic modeling is applied to identify the abstract "topics" that exist in a collection of documents. However, binary topic classification suffers when the dataset is imbalanced and has the effect of bias learning.

The results from the different combinations of pre-processing tasks have been obtained and analyzed by Uysal and Gunal (Uysal & Gunal, 2014). They have empirically examined the effects of applying various pre-processing tasks and demonstrated the importance of applying data pre-processing tasks. Interestingly, their results vary significantly across a different combination of pre-processing tasks on Turkish and English news datasets. The pre-processing tasks used in the experiment are 16 different combinations of tokenization, stop-word removal, lowercase conversion, and stemming on e-mails and news datasets. Their empirical results demonstrated the importance of sequence and combination of pre-processing tasks, as the best combination is able to perform 517% better than the worst combination on average in terms of F1-score. Nonetheless, as different techniques have their own preferred method of pre-processing, the proposed method is not suitable as it requires one to evaluate a different combination of pre-processing techniques for specific domains experimentally.

In the field of text classification, Howard and Ruder proposed Universal Language Model Fine-tuning (ULMFiT), a method for transfer learning and fine-tuning (Howard & Ruder, 2018). Their study has catalyzed the development of transfer learning in text classification tasks for NLP, especially when a limited amount of labeled data is presented to the model for training. This study has proved the importance of fine-tuning, yet their proposed fine-tune process is highly dependent on the model and datasets. When a model is trained with a different dataset, the learning performance of the model differs as the model might learn specific data better. Thus, the fine-tuning process should be tailored for each model and dataset. Zschech et al. proposed the design of a system that provides intelligent assistance, which can offer the most suitable suggestion of class according to the input (Zschech et al., 2020). Their classification models primarily focus on SVM and KNN. Their system takes the user's problem statement as input and pre-processes the input with NLP before applying classification. However, the complexity of this system design introduces large execution overhead as a long time for training necessities and needs a specialized processor such as GPU to execute.

Krishnalal et al. designed an intelligent system for online news classification based on a hybrid model that includes SVM and Hidden Markov Model (HMM) (Krishnalal et al.,2010). Once the text has been pre-processed, feature selection is applied to the text using HMM. Then the text is classified based on the SVM model. Yet, a different dataset domain can be used in the experiment to show the robustness of the model for prediction in various areas of interest. A hybrid text classification approach which is based on K-nearest neighbor (KNN) and SVM is implemented by (Wan et al., 2012). The parameters of a model have a high impact on the model's performance, leading to either positive or negative effects. Their proposed work has managed to overcome this problem by lowering the dependency of parameters on the classifiers. However, the running time of this approach is high due to degradation in its efficiency of the use of the validation set.

*2.2 Machine Learning Techniques*

Several different machine learning algorithms have been proposed to build the model for text classification. Aurangzeb et al. reviewed the methodologies and essential techniques for different machine learning algorithms to solve the tasks (Aurangzeb et al., 2010). More studies have been conducted to explore new strategies that can improve the efficiency throughout the classification process in different specialized fields. We briefly review the most recent work in this area.

*2.2.1 Support Vector Machine*

Srivastava and Bhambhu explore Support Vector Machine (SVM) for text classification (Srivastava & Bhambhu, 2010). Different kernel functions have been applied on different datasets, in which the results are compared and analyzed. However, they did not consider any pre-processing tasks. Dadgar et al. proposed a text mining approach using TF-IDF and SVM for news classification. The proposed approach consists of three main steps, text pre-processing, feature extraction using TF-IDF, and classification based on SVM (Dadgar et al., 2016). Nonetheless, the performance is only made on the trained model, and thus a bias judgment might be made during the training and overfits the model. Islam et al. also applied TF-IDF with SVM for the Bengali document classification task (Islam et al., 2017). They demonstrated the robustness of their implemented text classification model in different grammar languages, such as Bengali, other than English. The model achieved an average accuracy of 92.57%, as the model is trained with a large-scale dataset, and the performance using a smaller dataset might be affected negatively.

*2.2.2 Deep Learning*

Convolution networks (ConvNets) is a deep learning algorithm that is designed to take in an input image, then assign weights and bias to aspects in the image. The objective of ConvNets is to differentiate between different images. As ConvNets will require a large dataset to work, it is hardly applied to text classification tasks. Zhang et al. applied character-level ConvNext for text classification (Zhang et al., 2015). The performance of the model might be degraded due to the lack of training by using a smaller-scale dataset.

A new architecture based on Very Deep Convolutional Networks (VD-CNN) is introduced by (Conneau et al., 2017) for text processing. This deep problem-specific architecture can obtain a better performance than a one hidden layer neural network that is fully connected as it can develop hierarchical representation. However, the increase of depth in the convolutional network has also increased the complexity of the model with the complicated function while increasing the probability of overfitting and affecting the network's generalization.

Yin et al. have conducted a study comparing CNN and RNN for NLP (Yin et al., 2017). Based on the result, RNN has the best and robust performance for a broad range of tasks. Optimization of parameters tuning is shown to be important as it will affect the performance of the model directly. However, training an RNN is a difficult task due to its architecture. RNN model is not a feed-forward neural network, and it has a more complex signal movement. The RNN model also faces a vanishing gradient problem as the model is trained by the backpropagation method. The gradient tends to vanish over time when it is passed back through many time steps.

Moreover, Yao et al. have presented the utilization of graph convolutional networks for text classification (Yao et al., 2019). A single text graph is built for a dataset according to the document word relations and the word co-occurrence. Then, the Text Graph Convolutional Network (Text GCN) is applied to learn from the dataset. However, the GCN mode is inherently transductive (transductive learning does not build a predictive model, also known as semi-supervised learning), as unlabeled test document nodes are included in the GCN training. This is a major limitation because the model could not compute embeddings in a reasonable amount of time. It will not be able to predict unseen test documents accurately.

*2.3 Fine-tuning Methods*

The use of BERT in document classification is studied by Adhikari et al. to accept the hypothesis that BERT can be fine-tuned to achieve state-of-the-art results using several datasets (Adhikari et al.,2019). The fine-tuning of BERT is done by optimizing the number of epochs, batch size, learning rate, and some other hyperparameters. The authors have analyzed the model performance with different parameter values and found the most suitable parameters. However, the parameter values vary across different datasets, and there is a need to find the optimal value for the different datasets. Nonetheless, this work has become one of the foundations in studying fine-tuning of BERT for document classification. Zhu et al. made comparisons between BERT and SVM to identify and categorize hate speech in social media (Zhu et al., 2019). A pre-trained BERT Transformer is used, and the authors have further fine-tuned it. This study has shown the capabilities of the BERT model in solving text classification tasks, but the fine-tuning technique used by the authors only involved the addition of an extra layer on top of the base model. The authors implement no further improvement of the pre-trained BERT model in their work.

Sun et al. have investigated several fine-tuning methods for BERT by conducting experiments on text classification tasks (Sun et al., 2019). They proposed a general solution consisting of three main methods: pre-train BERT with in-domain data and within-task training data, fine-tuning strategies, and multi-task fine-tuning. But, training the model with more data will also increase the computational time especially when the machine learning model is only targeted at a specific domain. Zhang et al. reviewed standard practices in fine-tuning BERT (Zhang et al., 2020). Based on the review, many researchers have demonstrated that fine-tuned BERT gives a superior and state-of-the-art performance in the NLP area. Common practice typically focuses on three optimization methods: 1) bias correction in BERTAdam optimizer, 2) re-initialize the top later of pre-trained BERT, and 3) allocation of additional training time for stabilizing the fine-tuning algorithm. However, these approaches are not suitable for smaller datasets due to insufficient training data after the layers are re-initialized. This has an effect on deteriorating the model's performance.

A new mechanism is introduced by Xu, Qiu, Zhou, and Huang that improves the fine-tuning of BERT by self-ensemble and self-distillation methods (Xu et al., 2020). Initially, a few base models are combined by using the average parameter value. Then, knowledge distillation is carried out by allowing the current BERT model to learn from the self-ensemble model. However, this algorithm involves the pre-training process of BERT and high training time that can reach up to 54 hours using TPUv2. An optimization algorithm is required to reduce the overall complexity and training time in order to make their methods practical. As for most machine learning models, the performance of the models will show a drastic deterioration when there is a change in domain and training data. Ma et al. have presented a novel framework that aims to improve model adaptability when applied to different domains (Ma et al., 2019). BERT-based domain classifier is first trained with different data that are labeled with their respective domain. However, this framework is also not useful when the main objective of the model is focusing only on a specific domain, and further, the approach also requires a large number of training data. Other than focusing mainly on the pre-trained BERT model, researchers have also been looking at combining the advantages of BERT and other classic machine learning algorithms, such as CNN, RNN, SVM, and others. Zheng and Yang proposed a new method by using a BERT-CNN model to improve the performance of BERT in text classification (Zheng & Yang, 2019). In this model, CNN is used to transform the task-specific layer of BERT to acquire the text's local representation. Similarly, this proposed method requires higher computational time due to the computational burden of the CNN classifier in the training process.

In the study by Sajjad et al., the performance of BERT, RoBERTa and XLNet models are able to maintain up to 98% of their original performance when 40% of their architecture layer are pruned (Sajjad et al., 2021). In this work, the authors have shown that dropping layers in pre-trained models can lead to efficient transfer learning while providing a cheap way to get a smaller model of BERT rapidly in terms of memory and speed. Although the proposed models perform more efficiently, their accuracy of prediction results does not improve compared to the original models. Instead, we proposed a new method in our work, including pruning, while our models achieved higher accuracy than the original model.

**3 System Structure**

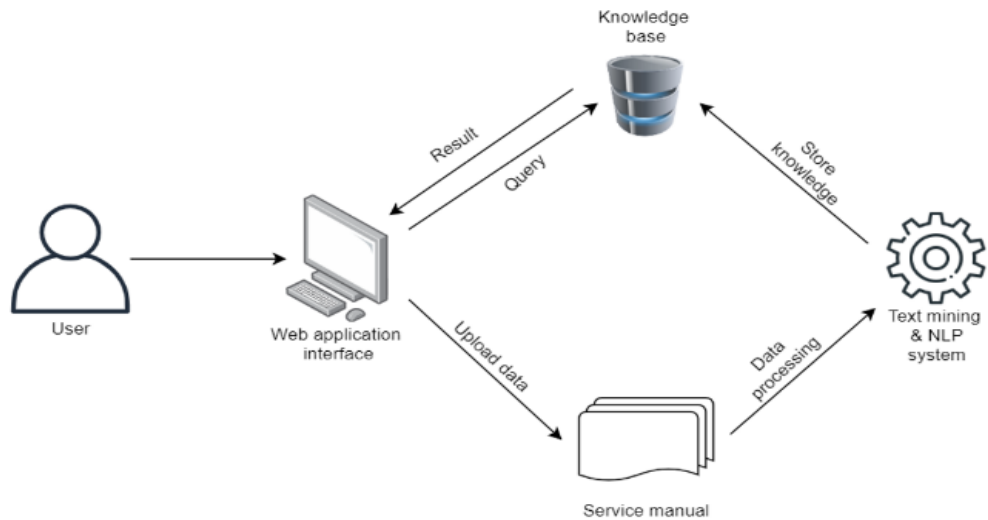

**Figure 1** System Architecture

The proposed system structure has been shown in Figure 1. Once the web application is started, the user will have two options: uploading the service manual or sending a search query. When the user uploads a service manual to the web application, data extraction and data classification processes will be carried out automatically. As for the query function, if the user decides to send a search query on the web application, the requested data will be displayed on the web application. The fine-tuning process is shown in Figure 2(a). Data extraction will be carried out during the initial stage once the service manual is successfully uploaded to the system. All the text data will be read and extracted for the upcoming process to mine the knowledge. Before proceeding with text classification, the data will need to be pre-processed as the unstructured data might consist of useful data and useless data and noise, which will affect the performance of classifiers (Kuma et al., 2018). The useless data does not contain any beneficial information, such as duplicated data and irrelevant data captured due to the error during data extraction. The process and techniques used in pre-processing can be seen in Figure 2(b). The embeddings are added to the input to tokenize each word in the sentence.

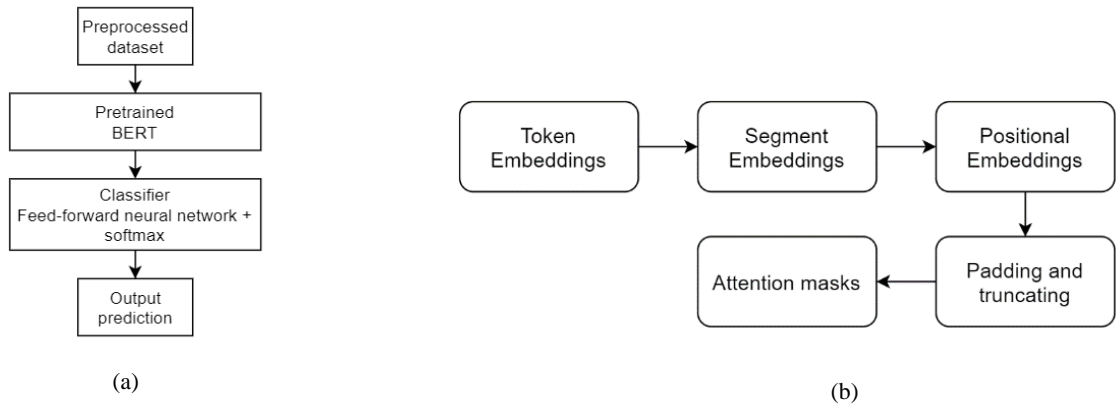

**Figure 2:** (a) Fine-tuning process (b) Pre-processing process

Segment embeddings and positional embeddings are also added to the input to differentiate between different sentences and indicate the position for each token from the original input sentence. As the proposed machine learning model, BERT has a limitation on the word limit, which is 512 words for the input. The padding and truncating processes are applied to ensure all input sentences in the dataset have a fixed length. In the experiment, the maximum length of the sentence is fixed at 64 since it is slightly larger than the maximum sentence length of the dataset used. Then, the attention mask for each token is stored as identification of the type of token. During pre-processing, the raw data is converted into a specific structure and format that can be fed directly to the machine learning model to predict the classes. The raw data are processed to the dimensionality of three: the input IDs, attention masks, and labels of each respective input sentence.

Once the raw data is processed, it proceeds to classify the text using the BERT model. The encoder receives input to the model in a sequence of the token after the data is pre-processed. The tokens are then converted into vectors and pass to the neural network for processing. This pre-trained BERT model is fine-tuned using our proposed fine-tune technique and is further trained with the provided dataset to generate a prediction of the data according to their related category.

During fine-tuning, an additional layer of the classifier with feed-forward neural network and softmax function is added on top of the pre-trained BERT model to further train the model according to the given task. The additional layer consists of untrained neurons and could be trained for classification tasks instead of training a model from scratch. Users may provide feedback regarding the classification result to improve the system's accuracy. The processed data will be stored in the knowledgebase according to the label of its classification as "knowledge". The user can use a query to fetch the "knowledge" out from the database more efficiently.

**4 Methods**

*4.1 Data Extraction*

The structure of service manuals varies across different brands or models, and the information from the service manuals forms "knowledge" that the user needs. Data extraction is required for the system to extract the wanted data from the document and transform it to the correct format that the proposed classification model can process. Physical extraction is used in this system as it reads the raw data from the input document, then converts the format of data for output.

In this study, tabula-py is used to read the table data from the input document in PDF format. Then, the raw data is converted to TSV, which can directly pass to the classification model as input. As the input document might contain unwanted data, such as the figure's caption, table title, intervention by domain expert is needed here to check and review the extracted data. This review process is required to ensure the system could perform well in classifying the data in the specific domain while cleaning the extracted data by removing the unwanted information.

*4.2 Multi-model Training, Pruning and Validation*

The random ordering of training data will affect the performance of the BERT model after fine-tuning (Sajjad et al., 2021). This problem often negatively affects the model performance, primarily when a smaller dataset is used to fine-tune the BERT model. When the last set of training samples mainly consists of noise, it will lead to backward training of the model as it is not learning correctly.

Therefore, multiple BERT models are initialized and trained during the fine-tuning process. The performance of each model is evaluated, and the potential best model is selected during this stage. Early stopping is carried out on the other models that did not perform well to minimize the training and computational time. Identifying the potential best model earlier during training is crucial since only one model is picked out among the others. All resources could be allocated to train the model further. The evaluation of the model during the selection process could also be used to indicate that the model has learned from the training data, which has a better order.

BERT is pre-trained on a large corpus consisting of data that might not be helpful and relevant when the BERT model will be used on a specific domain area. When a smaller dataset is used to fine-tune the BERT model, the model is highly prone to overfitting as it will rely more on the pre-trained weights. Pruning is introduced in this proposed method to allow room for new growth in the model's learning process while reducing the model capacity. The attention heads of the first layer are pruned to remove the information gained from pre-training. Since the model is undergoing the fine-tuning process, it is trained from scratch with the given dataset and will perform better in the respective task and domain. The first layer is chosen here as it consists of the least transferable knowledge and would not affect the model's performance negatively due to the lack of training data.

While tuning the hyperparameters of a model, the validation dataset is the sample of data that could be used to provide the fitness evaluation of the model that is unbiased on the training dataset. Although the model does not learn directly during validation, the model improves indirectly by using the validation set results to update the higher level hyperparameters. The number of validations is increased in this study as it could help in improving the accuracy of the model and its performance. Instead of evaluating the model once per epoch, the model is evaluated with the validation set 10 times during an epoch. This is to maintain the balance between the model performance and computational time, and it is a resource-intensive process.

*4.3 Fine-Tuning*

The pre-trained BERT model has already been encoded with the information learned from the huge corpus regarding the English language (Devlin et al., 2018). Hence, fine-tuning the model according to the given task and respective dataset is an effective method since less training time is required to complete this process. This model's pre-trained weights have also allowed the model to use a much smaller dataset to achieve a reasonable accuracy instead of building a new model from scratch.

An additional linear classification layer is added to the pre-trained BERT model to fine-tune it for the classification task. The library package from Hugging Face is used in this study as it provides an interface in PyTorch that could be used for modification of BERT with pre-built classes. Unlike training from scratch, fine-tuning the pre-trained model requires significantly less computational time and less data for further training as the pre-trained model has already trained on a large corpus. Only a few epochs are needed to train the model, and it will be able to achieve state-of-the-art results.

As suggested by Kumar et al., the parameter values are selected based on the author's recommended values (Kumar et al., 2018). A few experiments are carried out to identify the optimum value among the range of suggested values for each parameter of the BERT model. The proposed parameter values used here are shown in Table 2. The parameter values are typically tailored for all machine learning models to obtain the best model performance in their respective tasks and domain. In this study, BERT performs classification tasks by utilizing the machine learning algorithm to learn from experience and the given data. The BERT model is trained on a dataset that contains the data related to the given task domain and improves its performance from the past experience without being explicitly programmed. These parameters in the BERT model act as a balance between the overfitting and underfitting of the model during fine-tuning and tuning process and could be tweaked to achieve the best result. Different parameter values have been extensively tested on the BERT model in our work, and the optimal values for each variable are identified and used during the experiment.

*4.4 Web Application Knowledgebase*

Once the machine learning model has completed training, the entire model package is saved and downloaded for deployment purposes. A simple web application that provides CRUD (Create, Read, Update, Delete) operation is created using HTML, CSS, and Flask. This web application is connected to SQLite used as the database here to store the records as "knowledge". Once the model has completed training and has been fine-tuned, this model is integrated into the knowledgebase as an engine that will automate the data classification process according to the predefined category as in the trained data. Then, the processed data is stored in this knowledge base with their label accordingly. Users could proceed to perform query operations to retrieve the data as per their request.

*4.5 Model Descriptions*

*4.5.1 Proposed Fine-tuned BERT (Model 1)*

| **Algorithm 1** Model 1 |
|---|
| **Input**: Number of epoch $E$ |
| Set $E = 3$ |
| Preprocess data |
| Splitting data |
| Sampling data |
| Initialize BERT model |
| Prune first layer of model |
| **for** $e = 0$ **to** $E$ **do** |
|   Model training |
|   Model evaluation |
| **end for** |

The proposed algorithm to fine-tune the BERT model is shown in Algorithm 1. In this proposed method, only one model is focused on and used for training. Initially, the input data is pre-processed according to the steps shown in Figure 2(b). The dataset is split into 80% for training data while 20% is validation data for model evaluation. Data sampling is performed on both training and validation data randomly. As the data is sampled randomly without replacement, the process is carried out on a shuffled data. Hence, it would produce a different sequence of data that would be used to train the model every time data is sampled.

The model is initialized by loading the pre-trained BERT model from the Hugging Face library with the additional layer on top. The BERT model used is BERT-Base, Uncased, which has 12-layer, 768-hidden, 12-heads, and 110M parameters. The parameter values of the model are also initialized during this step to fine-tune the model. Before the model training is started, all attention heads in the first layer of the model are pruned. Then, the model is trained and evaluated for three epochs. The model's hyperparameters are updated automatically during each epoch through backpropagation to improve the model performance.

*4.5.2 Proposed Fine-tuned Multi-model (Model 2)*

The proposed algorithm to fine-tune the Multi-model BERT model is shown in Algorithm 2. As opposed to Model 1, three models are trained initially, and the best model among them is selected as Model 2. The input data is pre-processed according to the steps shown in Figure 2(b). The dataset is split into 80% for training data while 20% is validation data for evaluation. The process flow of fine-tuning for Model 2 is shown in Figure 3. Three models are involved in the fine-tuning process of Model 2, and three different sets of training data are used to train each model respectively. Then, each model undergoes the same series of steps, which are pruning, training and validation. After the first iteration, the best model is selected based on accuracy and continues to train for two epochs more before the completion of fine-tuning.

---

**Algorithm 2** Model 2

---

**Input**: Number of epoch $E$
Set $E = 3$
Preprocess data
Splitting data
Sampling data
Initialize three BERT models
**for** $e = 0$ **to** $E$ **do**
  if (e **equals** 0)
    Prune first layer of model
    Model training
    Model evaluation
    Select the highest accuracy model
    Continue to next iteration
  **end if**
**end for**

---

Data sampling is performed three separate times on both training and validation data randomly to produce three sets of sampled data. As the data is sampled randomly without replacement, the process is carried out on a shuffled data. Hence, the three sampled data have a different sequence of data that is used to train each of the three models. Three models are initialized by loading the pre-trained BERT model from the Hugging Face library with the additional layer on top for every three models. The BERT model used is BERT-Base, Uncased, which has 12-layer, 768-hidden, 12-heads, and 110M parameters. Same parameter values are initialized for each of the models to start learning. During the first epoch, all attention heads in the first layer of the three models are pruned. Then, the three models are trained and evaluated based on the validation data. According to the evaluation result, the model with the highest accuracy value is selected to be the best model. The best model is further trained and evaluated for two more epochs.

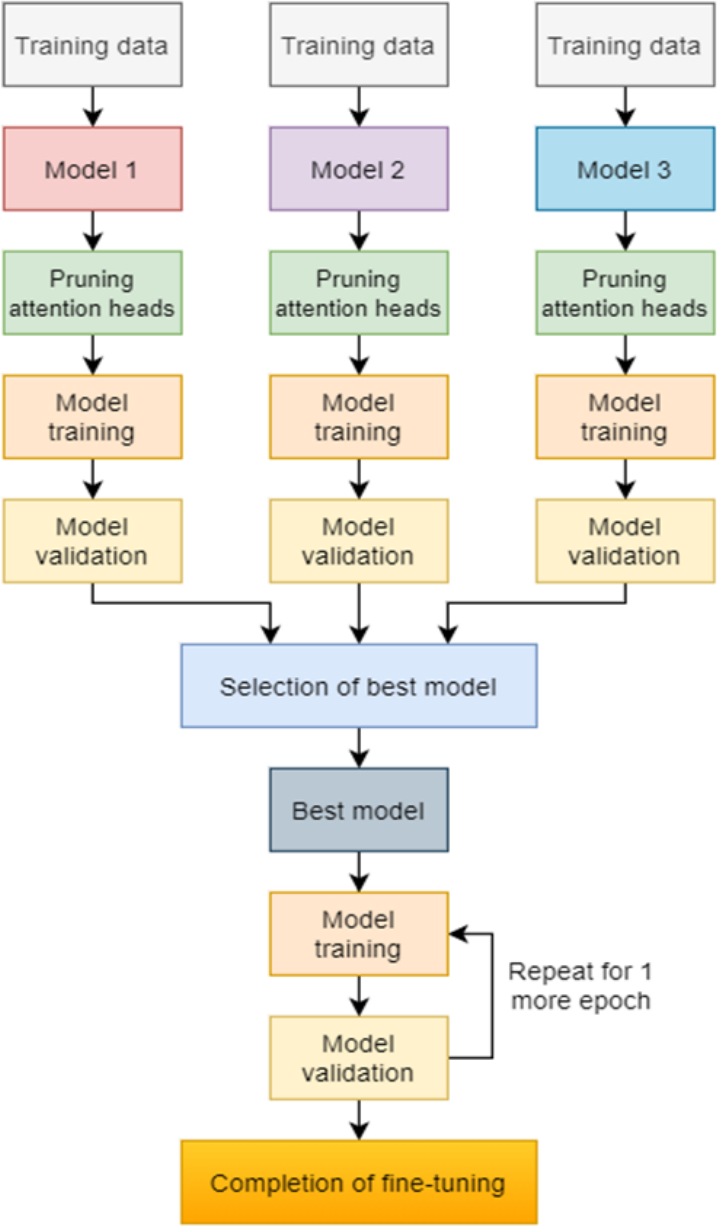

**Figure 3:** Fine-tuning process for Model 2

## 5 Implementation

### 5.1 Dataset

**Table 1:** Distribution of instances in training and test sets

| Classes | No. of Instances | |
| --- | --- | --- |
| | Training | Test |
| 0-unacceptable | 2528 | 162 |
| 1-acceptable | 6023 | 354 |
| Total | 8551 | 516 |

The dataset used in this paper is Corpus of Linguistic Acceptability (CoLA) and consists of 10,657 sentences from 23 linguistics publications (Warstadt et al., 2019). Although the target application of work is text classification for service manual documents, no open-source service manual dataset is currently available. Nonetheless, the CoLA dataset consists of the different categories of sentences extracted publication, which is highly similar to data types of the service manual. Hence, it is chosen to benchmark our approach in data classification using the BERT model. Each sentence is annotated for its acceptability in terms of grammatical by their original authors expertly. The dataset is split into two parts, which are used for training and testing. There are four attributes and 8551 instances in the training set, while 516 instances in the testing set. "Attribute 1" contains the code representing the source of the sentence. "Attribute 2" and "Attribute 3" represent the acceptability judgment label and the acceptability judgment as notated initially by the author, respectively. Finally, "Attribute 4" contains the sentence itself. The target class has two values: 0 for "unacceptable" and 1 for "acceptable". The number of instances for the training and testing set is shown in Table 1. There is no missing data in the dataset. However, the dataset is imbalanced due to the difference in the number of instances between the two classes.

*5.2 Experiment Setup*

Table 2: Parameters used for model training and tuning

| Parameter Name | Value |
|---|---|
| No. of Epochs | 3 |
| Batch Size | 16 |
| Optimizer | AdamW |
| Epsilon | 1e-8 |
| Learning Rate | 2e-5 |
| Training Steps | 1284 |
| Training: Testing | 80:20 |

The experiment of this study is carried out using the hardware by Google Colab GPU, Tesla T4. Once the dataset is loaded, the raw data are pre-processed using token embeddings, segment embeddings, and positional embeddings. These steps are needed to prepare the data to train two different NLP tasks: masked language modeling and sentence prediction. Once the input data is processed to the proper format, the BERT model can proceed to be explicitly fine-tuned according to the given task while achieving high performance. In this study, BertForSequenceClassification is integrated with "bert-base-uncased" as the BERT mode. BertForSequenceClassification is an interface provided by the Hugging Face PyTorch implementation built on top of a pre-trained BERT model. It is designed to handle a specific task, which is to classify sentences in this case. AdamW is used as the optimizer for this model. The parameters are tuned according to Table 2. The model is further trained with the processed dataset to be well-suited for the given task. Four BERT-based models are tested in this experiment: BERT (Devlin et al., 2018), BERT-CNN (Safaya et al., 2020), and the two proposed models – Model 1 and Model 2. The initialization for Model 1 and Model 2 is done by loading BertForSequenceClassification as the model. Each model is then further trained with the given dataset to learn from the experience and improve their performance in classifying the input data.

## 6 Results and Discussion

The performance of the proposed Multi-model Transfer Learning approach has been evaluated based on accuracy, recall, precision, F-measure, Matthews correlation coefficient (MCC), and ROC area. Tab. 3 shows the confusion matrix of all models. It can be seen that Model 2 is able to classify the text with the most true-labeled text of 429 instances consisting of true positive (TP) and true negative (TN) counts, while the base BERT model only has 417 cases of them.

**Table 3:** Parameters used for model training and tuning

| Model | TP | FP | TN | FN |
|---|---|---|---|---|
| BERT | 333 | 78 | 84 | 21 |
| BERT-CNN | 334 | 83 | 76 | 20 |
| Model 1 | 336 | 72 | 90 | 18 |
| Model 2 | 339 | 72 | 90 | 14 |

As shown in Table 4, both Model 1 and Model 2 are able to outperform the base BERT and BER-CNN models in terms of MCC score and accuracy. Model 1 and Model 2 have successfully enhanced the model's performance for MCC results from 0.529 to 0.591 and 0.592 respectively, in contrast to the BERT model. There is an increase of 11.72% and 11.91% for Model 1 and Model 2 from the base BERT model. While comparing the performance with BERT-CNN, Model 1 achieves 16.34% higher accuracy while Model 2 outperforms with 16.54% higher accuracy. As a high value of F1-score is desirable for adequate classification performance, Model 1 and Model 2 both have achieved higher F1-score and ROC area as well than the base BERT model. Therefore, it can be concluded that both the proposed BERT models have succeeded in outperforming the base BERT model. Despite the base BERT model has the lowest training time among all the models in this experiment, the significant performance boost of our proposed methods outweighs the slightly increased training time.

Table 4: Comparison of performance measures

| Metric | BERT | BERT-CNN | Model 1 | Model 2 |
|---|---|---|---|---|
| MCC | 0.529 | 0.508 | 0.591 | 0.592 |
| Accuracy | 0.808 | 0.800 | 0.831 | 0.831 |
| Precision | 0.810 | 0.801 | 0.828 | 0.825 |
| Recall | 0.941 | 0.944 | 0.952 | 0.958 |
| F1-Score | 0.871 | 0.866 | 0.886 | 0.886 |
| ROC Area | 0.730 | 0.716 | 0.760 | 0.757 |
| Training Time (s) | 430 | 581 | 546 | 840 |

The experimental results for selecting the optimum parameter values can be obtained from Tables 5 – 7. As seen in the results, not all models were able to perform well when using the suggested parameter values from Table 2. Each model has its own set of parameter values tailored and tuned accordingly to ensure the best performance on the respective task. Meanwhile, both proposed models, Model 1 and Model 2 are shown to have the best performance by using the parameter values in Table 2.

Table 5: Comparison of MCC with varying batch size

| Batch Size | BERT | BERT-CNN | Model 1 | Model 2 |
|---|---|---|---|---|
| 4 | 0.529 | 0.503 | **0.530** | 0.460 |
| 8 | 0.546 | 0.503 | **0.550** | 0.504 |
| 16 | 0.529 | 0.519 | 0.591 | **0.592** |
| 32 | 0.550 | 0.519 | 0.556 | **0.561** |
| 64 | 0.519 | 0.540 | **0.550** | 0.529 |

**Table 6:** Comparison of MCC with varying epoch

| Epoch | BERT | BERT-CNN | Model 1 | Model 2 |
|---|---|---|---|---|
| 2 | **0.540** | 0.477 | 0.535 | **0.540** |
| 3 | 0.535 | 0.514 | 0.591 | **0.592** |
| 4 | 0.529 | 0.503 | 0.581 | **0.586** |

**Table 7:** Comparison of MCC with varying learning rate

| Learning Rate | BERT | BERT-CNN | Model 1 | Model 2 |
|---|---|---|---|---|
| 1e-5 | **0.545** | 0.519 | 0.529 | **0.545** |
| 2e-5 | 0.529 | 0.503 | 0.591 | **0.592** |
| 3e-5 | **0.561** | 0.498 | 0.545 | 0.545 |
| 4e-5 | 0.505 | 0.536 | 0.519 | **0.545** |
| 5e-5 | 0.534 | 0.541 | **0.546** | 0.530 |

In Table 8 and Table 9, the comparison results of the proposed methods are provided by varying the number of in-training validation and pruning. The default value of in-training validation is set to 10. Both models are shown to be performed better when in-training validation and pruning are incorporated in the models.

**Table 8:** Comparison of MCC for in-training validation and pruning

| Model | Model 1 | Model 2 |
|---|---|---|
| Without in-training validation | 0.529 | **0.545** |
| With in-training validation | 0.591 | **0.592** |
| Without pruning | 0.545 | 0.545 |
| With pruning | 0.519 | **0.545** |

**Table 9:** Comparison of MCC with varying in-training validation

| Varying in-training validation | Model 1 | Model 2 |
| --- | --- | --- |
| 2 | 0.581 | 0.540 |
| 5 | 0.560 | 0.586 |
| 10 | 0.591 | 0.592 |
| 12 (35) | 0.581 | 0.556 |
| 15 (28) | 0.560 | 0.550 |
| 20 | 0.535 | 0.540 |

Due to the imbalance ratio in the dataset, the MCC score is the most appropriate metric to evaluate the performance of each model. We can observe that our two proposed BERT models have shown their ability to classify text according to the predefined category. Both proposed BERT models have achieved higher performance, especially in MCC scores. We can also observe that Model 1 has better stability than Model 2. Model 2 is unstable due to the random seed used in the experiment. The seed number of the model is not fixed as the model is driven by the randomness of data order since it has shown that certain data order has performed well than others. If the randomness of the model is overcome by using a fixed seed number, the model will not be able to identify the best training data order as a fixed order will be used. Model 1 can generate a stabilized result due to a fixed seed number in the model to generate a reciprocate result. This model's training time is also lower than Model 2 as only one model is used for training. Although the effectiveness of training data order is not examined in this model, this model managed to pull off a competitive result compare to Model 2 while offering a lower computational time. In summary, the experimental results have shown the potential of our two proposed methods in improving the BERT model for text classification.

**7 Conclusion**

Automated Document Knowledge Mining and Classification System is valuable for automating categorization and data storage in the knowledge base without human intervention. Users can easily retrieve the desired knowledge from the knowledge base with a minimal amount of time. As evidenced by our experimental studies, both proposed BERT models, namely Model 1 and Model 2, outperformed the BERT and BERT-CNN models. Our models achieve superior performance by applying multi-model transfer learning with pruning and fine-tuning techniques. Further work could be done to improve the classification accuracy of the BERT model by training it on a specific task or domain. It is also possible to implement automatic noise removal for raw data and to include it in the model as part of ongoing research. Furthermore, the optimization algorithm could be integrated into the fine-tuning of the model to lower the computational time. More features could also be added to the web application while improving the interface so the system will be ready for production.

**Conflicts of Interest:** The authors declare that they do not have any conflicts of interest regarding the present study.